\documentclass{svproc}

\usepackage{url}
\usepackage{cite}
\usepackage{amsmath,amssymb,amsfonts}
\usepackage{algorithmic}
\usepackage{graphicx}
\usepackage{textcomp}
\usepackage{xcolor}
\usepackage{cite}
\usepackage{amsmath,amssymb,amsfonts}
\usepackage{algorithmic}
\usepackage{graphicx}
\usepackage{textcomp}
\usepackage{xcolor}
\usepackage{tabularx}
\graphicspath{{images/}}

\graphicspath{{./}}

\begin{document}
	
	\mainmatter              % start of a contribution
	\title{An IoT Based Framework For Activity Recognition Using Deep Learning Technique\\
	}
	\titlerunning{Activity Recognition}
	\author{Ashwin Geet D'Sa \and Dr. BG Prasad
	}
	
	\institute{Department of Computer Science,\\
		B.M.S. College of Engineering, Bengaluru, India,\\
		\email{win.12894@gmail.com},\\
		\and
		\email{drbgprasad@gmail.com}
	}
	
	\maketitle

\begin{abstract}
 Activity recognition is the ability to identify and recognize the action or goals of the agent. The agent can be any object or entity that performs action that has end goals. The agents can be a single agent performing the action or group of agents performing the actions or having some interaction. Human activity recognition has gained popularity due to its demands in many practical applications such as entertainment, healthcare, simulations and surveillance systems. Vision based activity recognition is gaining advantage as it does not require any human intervention or physical contact with humans. Moreover, there are set of cameras that are networked with the intention to track and recognize the activities of the agent. Traditional applications that were required to track or recognize human activities made use of wearable devices. However, such applications require physical contact of the person. To overcome such challenges, vision based activity recognition system can be used, which uses a camera to record the video and a processor that performs the task of recognition. The work is implemented in two stages. In the first stage, an approach for the Implementation of Activity recognition is proposed using background subtraction of images, followed by 3D- Convolutional Neural Networks. The impact of using Background subtraction prior to 3D-Convolutional Neural Networks has been reported. In the second stage, the work is further extended and implemented on Raspberry Pi, that can be used to record a stream of video, followed by recognizing the activity that was involved in the video. Thus, a proof-of-concept for activity recognition using small, IoT based device, is provided, which can enhance the system and extend its applications in various forms like, increase in portability, networking, and other capabilities of the device. The source-code for this work has been made available\footnote{https://github.com/GeetDsa/Activity\_Recognition\_3D\_CNN}.\\
\end{abstract}

\section{Introduction}
\subsection{Activity Recognition}
Activity recognition is the ability to identify and recognize the action or goals of the agent, the agent here can be any object or entity that performs action, which has end goals. The agents can be a single agent performing the action or group of agents performing the actions or having some interaction. One such example of the agent is human itself, and recognizing the activity of the humans can be called as Human Activity Recognition (HAR)\cite{aggarwal_human_2011}. In the last few years, automatic recognition of human activities has gained much attention in the field of vision based technologies due to its increasing demands in practical applications, such as surveillance environments, healthcare systems and entertainment environments. In a surveillance system, the automatic identification and classification of unusual and abnormal activities can be made. This would aid in alerting the concerned authority or the person monitoring the given environment, for example group attacks or fights can be recognized and the concerned authority can be informed about it. In systems belonging to entertainment environment, activity recognition can be used for Human Computer Interaction (HCI) systems, which would involve identifying the activity of the person and responding to the activity of the actor. For example, we can use this in simulation of the game, where the human remains one of the players and the computer responds to the action of human player by simulating the computer based animations. In a healthcare system, the recognition of activities can help in activities such as rehabilitation of patients, where the activities of the patients can be monitored in real-time.  Human activity recognition is not just limited to a few of these applications, but also being used in various other applications. The work is implemented in two stages. In the first stage, an approach for the Implementation of Activity recognition is proposed using background subtraction of images, followed by 3D- Convolutional Neural Networks. The impact of using Background subtraction prior to 3D-Convolutional Neural Networks has been reported. The system design, and the results for this section are discussed in section 3 and 4 of this article. In the second stage, the work is further extended and implemented on Raspberry Pi, that can be used to record a stream of video, followed by recognizing the activity that was involved in the video. Thus, a proof-of-concept for activity recognition using small, IoT based device, is provided, which can enhance the system and extend its applications in various forms like, increase in portability, networking, and other capabilities of the device. The architecture used for proof of concept of this extension is discussed in section 5 of this article.
\subsubsection{Background Subtraction}
Background subtraction is an image processing technique used for foreground detection in  videos, i.e, it is used for identifying the object in motion when there is a static background. The common approach for background subtraction is the use of static reference image with no moving objects, which can then be used to identify the moving objects in the foreground \cite{piccardi_background_2004}.
\subsubsection{Deep Learning}
Traditional machine learning algorithms require the pre-processing of the data followed by representation of the data prior to the use of machine learning algorithms \cite{bengio_representation_2013}. Representational learning is the class of machine learning that learns the features or the representations that can be effectively used  by the machine learning algorithms. These algorithms transform the data into representations. Deep learning can be best understood as the hierarchy of sequence of processing stages, where each layer or stage represents the given data in its own way\cite{bengio_learning_2009}. Each layer or stage transforms one representation obtained from previous stage to another representation, where the data is transformed from lower level features to higher level representation. The last layer of the deep learning architecture is used for the prediction tasks in the case of supervised learning. 
\subsection{3D-Convolutional Neural Network}
The idea behind convolutional neural networks is that, a fully connected feed-forward neural network can be used to simulate the learning of the features as well as perform the task of the classification \cite{aghdam_convolutional_2017}. The challenge of using the neural networks for the classification problems where the input data is an image, is the number of neurons that are used. Since each pixel may be connected to one neuron in the input layer, it increases the number of model parameters for the training. Convolutional Neural Networks is the way to reduce the number of model parameters for learning a model with the neural network architecture. This is done by using small sized filters, which is passed over the entire image, hence a neuron corresponds to the cell of a filter than to the cell of an image and the number of weights can be further reduced by sharing the same weight across all the cells in a single filter. Convolutional neural network replaces the problem specific handcrafted filters with that of trainable filters \cite{ji_3d_2013}. The convolutional neural networks are also known for their translational and scale invariance. The 2D convolutional neural networks learn the spatial features over the 2D space of the two-dimensional image, whereas the 3D Convolutional Neural Networks learn the spatio-temporal features over the sequence of 2D images, thus learning the additional temporal features using sequence of the images. Thus,2D-convolutional neural networks used for learning the spatial features of images can be extended to 3D convolutional neural networks to learn the spatial-temporal features for the videos \cite{baccouche_sequential_2011}.\\
\section{LITERATURE REVIEW }
\indent Survey on various methods of activity recognition and action recognition were made. Where, some works for action recognition were based on trajectory of the motion and few of them were based on pose estimation. The works that are used for activity recognition follow the steps in the sequence: 1. Segmentation of the video, where the region of interest or presence of humans is detected, 2. Feature Extraction, where the required features are extracted based on the motion or the pose of the humans. 3. Feature Representation, here the  extracted features are represented using the feature vectors or feature descriptors. In case of topic modeling, code-book is used to represent these features. Finally, training and testing is done using classification model. A detailed survey is reported in the our previous work, and briefed out in this paper. \cite{ashwinBGP} \\

\indent Segmentation in human based activity recognition acts like a preprocessing step and this may or may not be performed based on the steps used in feature extraction and feature representation. It is observed that some algorithms perform feature extraction without the use of segmentation. Segmentation is defined as dividing the entire image into group of subsets, typically one of the subset must contain the region of our study that has to be processed further. Pre-processing techniques such as ‘background subtraction’ or ‘foreground object extraction’ has been used for this purpose \cite{vishwakarma_survey_2013}. The other preprocessing techniques may involve marking of key start and end frame manually.\\
\indent Feature Extraction, is the step that involves extraction of features such as shape, silhouette, motion information, etc, that can be represented so that the classification algorithm may be applied over it. Feature extraction varies based on the type of approach that is used for activity recognition. Activity recognition can be achieved using two approaches, 1. Motion or Trajectory Based Approach- where the features represent the motion information of the humans or objects. This type of approach is used in few of the works \cite{boufama_trajectory-based_2017,abdelhedi_human_2016,guo_continuous_2014}  and 2. Pose Based Approach- where the pose of the human is considered and acts as feature for the action or activity recognition \cite{youssef_human_2013,jalal_depth_2012,huynh-the_interactive_2016}.\\
\indent The features used for motion based approach in few of the notable works are- Interest point (IP), Lucas-Kanade (LK) and Farnback (FB) trajectories \cite{boufama_trajectory-based_2017}, Optical flow \cite{guo_continuous_2014}, etc. The features used for pose based approach used in few of the notable works are- Human Joint-Coordinates along with distance and angle features representing joints, where the work \cite{huynh-the_interactive_2016} has performed 14-part and 26-part joint coordinates, silhouette extraction \cite{vishwakarma_proposed_2016}, fuzzy model for the selection of key pose \cite{vishwakarma_proposed_2016}, depth silhouettes \cite{jalal_depth_2012}, etc.\\
\indent The extracted features are then represented, so that the classification algorithms may be applied over them. Here the feature representation depends on the approach used, since the feature representation depends on the extracted features. The features can be represented after applying dimension reduction algorithms like principal component analysis (PCA), local linear embedding (LLE) or Linear Discriminant Analysis (LDA) \cite{vishwakarma_proposed_2016,jalal_depth_2012}.\\
The features extracted can be represented as a single descriptor or a topic model- where the set of words map to a particular topic. In the work \cite{huynh-the_interactive_2016}, Pachinco allocation model, a topic model is used for the feature representation, where the features are angle and distance parameters corresponding to the human pose. Other topic models used are bag of words algorithm and Latent Dirichlet Allocation, which requires the generation of code-book of words. Here the words are essentially derived per video frame, where the set of words map to particular poselet, which in turn may map to an action, and which finally maps to an activity. The other feature representation use ‘Radon-Factor’ or ‘R-descriptor’ obtained after applying radon-transform, here the  Radon filters are invariant to scaling of the shapes of the human pose, which are effective when the size of appearance of the person changes \cite{vishwakarma_proposed_2016}. Other ways of feature representation includes spatial distribution of edge gradient (SDEG) \cite{vishwakarma_proposed_2016}, Translation and Scale Invariant probabilistic Latent Semantic Analysis model (TSI-pLSA)\cite{guo_continuous_2014}. Hu moments and Zernike moments feature vector are used in the case of work \cite{abdelhedi_human_2016}, where the optical flow was the feature extracted.\\
\indent The classification algorithm is used to create the classification model based on the training data, where this created model is used to test the video for recognizing and classifying the activity. Few of the classification algorithms used for activity recognition are multi-class Support Vector Machine(SVM) classifier \cite{vishwakarma_proposed_2016,boufama_trajectory-based_2017}, Expectation Maximum (EM) Algorithm, Bayesian decision \cite{guo_continuous_2014}, Hidden Markov Models (HMMs)\cite{jalal_depth_2012}, Feed-forward neural networks \cite{abdelhedi_human_2016}, etc.\\
\indent It can also be observed that the concepts of Neural Networks and Deep Learning \cite{baccouche_sequential_2011,park_depth_2016,dobhal_human_2015,yu_fully_2017}, are used in recent approaches and networks such as Convolutional Neural Networks, Recurrent Neural Networks and LSTM’s are used. These types of neural networks have reduced amount of preprocessing, since CNNs are used to find the hidden patterns in the given data-set and also RNN takes time series data, which is very useful in gaining the temporal information.\\
\indent  The methodology used by various authors for the activity recognition along with the scope for future work as mentioned by the authors are discussed:\\
\\
\indent S. U. Park, et al.\cite{park_depth_2016}, proposed used of Recurrent Neural Network (RNN) for HAR. The joint angles are computed, and input feature matrix is created for the obtained joint angles, Recurrent Neural Networks are used for training the data. RNN consisted of 50 Long Short-Term Memory (LSTM)s with 90 hidden units, which prevented the vanishing gradient problem. Accuracy of  95.55\% was achieved using MSRC-12 data-set.\\
\indent Tushar Dobhal, et al.\cite{dobhal_human_2015}, proposed a method to classify the human actions by converting the 3D videos to 2D binary motion images. For the input video, background from each image is subtracted using Gaussian Mixture model. All the action sequence images are then combined to obtain a single image known as Binary Motion Image (BMI). Then, Convolutional Neural Networks(CNN) is used for learning, which does both extraction of features as well as classification. CNN requires less pre-processing compared to ANN. Accuracy of 100\% on Weizmann data-set and 98.5\% on MSR Action 3D data-set was achieved. The authors used MATLAB for extracting the BMI, and ConventJS to implement a 3 layer CNN.\\
\indent Sheng Yu, et al.\cite{yu_fully_2017}, proposed use of a two stream CNN in order to avoid the problem of overfitting in CNN and perform the action recognition. The input data is passed into two separate streams (Spatial and Temporal). The RGB video frames became the input to the spatial stream. Stacking of optical flow obtained using TLV1 method is used as input to the temporal stream. Learning rate of 0.00001 is used for 1st 10k iterations followed by 0.000001. Stochastic gradient descent is used for training the model. The streams of CNNs are treated as feature extractors and the last max-pooling layer is used as vector of features. Two fusion techniques are used to fuse the features: i) Linear weight fusion method is used to add the pixels of spatial and temporal feature maps where its weights signifies the importance; ii) Concatenation fusion, reshapes the combination of both the features into single vector. A Vector of locally aggregated descriptor (VLAD) and temporal pyramid pooling (TPP) are used together to obtain video level features. The classification is done based on SVM. Caffee toolbox is used to implement the CNN. Accuracy of 90.5\% on UCF101 data-set using linear weighted fusion technique and Accuracy of 63.4\% on HMDB51 data-set using linear weighted fusion technique has been reported.\\
\indent Thus, we can infer that there is no single straight forward method that can be employed for activity recognition. However, we have a choice of variety algorithms at every step that can be used for recognizing the activity, where few of the important steps include, feature extraction followed by feature representation, and then the classification over the represented feature, used to classify the activities. \\
Further, it can be observed that most of the IoT based activity recognition used sensor based activity recognition \cite{rodriguez2017iot,yao2018wits,castro2017wearable}. The sensors were either wearable or  embedded in mobile devices. However, through this work, vision based activity recognition is proposed which uses a static camera to record video, followed by processing the video using a pocket sized, portable computer, i.e, Raspberry-pi.

\section{System Design of Activity Recognition System}
\begin{figure}[htbp]
	\centerline{\includegraphics[height=100mm]{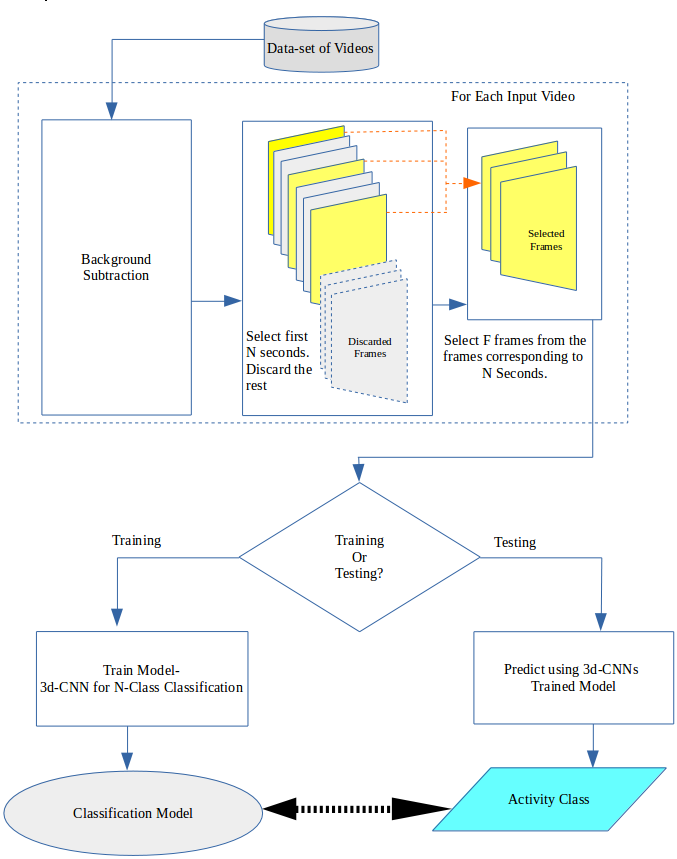}}
	\caption{System Architecture for Activity Recognition System}
	\label{SD}
\end{figure}

The system architecture for activity recognition system is shown in Fig \ref{SD}, depicting the overview of design adopted in this work.
\\
\textbf{Following is the brief description of the system design for activity recognition:}
\begin{itemize}
	\item The data-set of videos forms the input to the system which may reside on the storage device.
	
	\item The videos are split into 3 sets: training set, validation set and test set.
	\item For every video in the set, videos are first converted to gray-scale videos. Then, two copies of videos are made: i) one with background subtraction operation performed, and ii) without background subtraction operation.
	\item For each copy of the video, first ‘S’ seconds are considered. From each video, ‘N’ frames from the first ‘S’ seconds are considered by equal interleaving between the frames. This has been done to obtain equal sized inputs.
	\item Both the copies of videos (with and without background subtraction) are used separately to fit the 3D-Convolutional Neural Networks model. Training and validation set of videos are used for this purpose.
	\item For the purpose of evaluation of the trained model, prediction of the activity class and evaluation of accuracy is made using the test set of videos.
	\item For the purpose of fine tuning of model and improve the accuracy, data augmentation here is performed by flipping the image frames horizontally in the training set, followed by using the flipped images as part of training set. Also, flipping of the image frames of validation set is done and the videos with flipped image frames is concatenated with training set. This, increases the data in training set. Additionally by using only the videos with flipped image frame of validation set in the training set, ensure that same videos of validation set does not repeat in training set. The techniques of learning-rate decay
\end{itemize}

Three data-sets have been used for the purpose of implementing the activity recognition system and are explained below:

\textbf{KTH:} It consists of 6 classes- running, boxing, walking, jogging, hand waving, and hand clapping, it has been recorded in 4 environment, indoor, outdoor, with camera jitter and with noise. The actions are performed by single actor. The videos are black and white videos with the resolution of 160*120.

\textbf{Wiezmann:} It consists of 10 classes- jump, gallop sideways, walk, run, bend, one-hand wave, two-hands wave, jumping jack, jump in place, and skip. These actions are performed by 9 actors, hence the total count of 90 videos in the data-set. The videos are recorded by the static camera with the resolution of 180*144.

\textbf{UT-Interaction \cite{ryoo_ut-interaction_2010}:} It consists of 6 classes- hand shaking, hugging, kicking, pointing, punching, pushing. The data-set used for the study is the special set of UT- Interaction data-set where the frames are segmented out to consider only the actors in the video. Since the UT-interaction contains 2 sets, where one set is recorded with static background and no pedestrians in the videos, the other set being recorded with either the presence of the pedestrians or the noisy background. The segmented data-set considers the absence of pedestrians. Thus, UT-Interaction Segmented data set is considered for the study.\\

4 Models of CNNs are constructed for the purpose of study:\\
Model 1: 2 Layer 3D-CNN\\
Model 2: 3 Layer 3D-CNN without Dropout Layer\\
Model 3: 3 Layer 3D-CNN with Dropout Layer\\
Model 4: 3 Layer 3D-CNN with Dropout Layer and uses data-augmentation on Training set. This model also uses the technique of learning rate decay.\\
\\
Model 1, Model 2 and Model 3 are trained on adam optimizer \cite{kingma2014adam}. However, Model 4 has been trained on n-adam optimizer \cite{dozat2016incorporating}. \\
Dropout is a technique of regularization used to prevent over-fitting in CNN \cite{krizhevsky_imagenet_2017}and Neural networks\cite{srivastava_dropout:_2014}. The function of dropout layer is to avoid using certain neurons with probability `p' during the training phase.\\
Table \ref{ID} depict implementation details showing the parameters- the total number of videos considered for training, testing and validation, along with the number of frames `N' considered from the first `S' seconds of the video.
\begin{table}[ht]
\caption{Implementation Details}
\label{ID}
\begin{tabular}{|c|c|c|c|c|c|}
	\hline
	\textbf{Data-set} & \textbf{Train} & \textbf{Validation} & \textbf{Test} & \textbf{`S'} & \textbf{`N'}\\ \hline
	\textbf{KTH}&300&122&100&7&35 \\ \hline
	\textbf{Weizmann}&46&13&18&2&20 \\ \hline
	\textbf{UT Interaction}&300&122&100&7&35 \\ \hline
	
\end{tabular}

\end{table}
\section{Results And Discussion}
The test bench used for performing training and testing of activity recognition system had the following specification:\\
Operating System : Ubuntu 16.04 LTS\\
Memory(RAM)	  : 16 GB\\
Processor: Intel(R) Core(TM) i7-6500U CPU @ 2.50GHz\\
Graphic Memory: 4GB\\
API Used: Keras \cite{chollet2015keras} \\

\begin{figure}[htpb]
	\vspace{-1em}
	\centerline{\includegraphics[height=60mm]{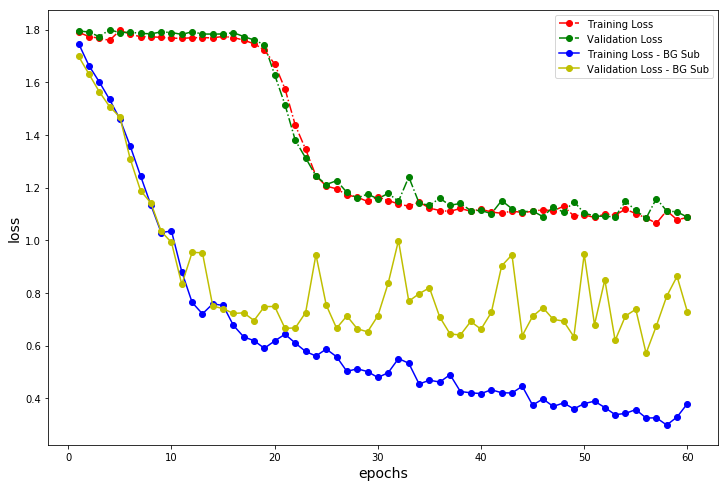}}
	\caption{Training and Validation loss for Model 1 using KTH Dataset}
	\label{KTH_M1}
	\vspace{-0.5em}
\end{figure}

\begin{figure}[htbp]
	\vspace{-1em}
	\centerline{\includegraphics[height=60mm]{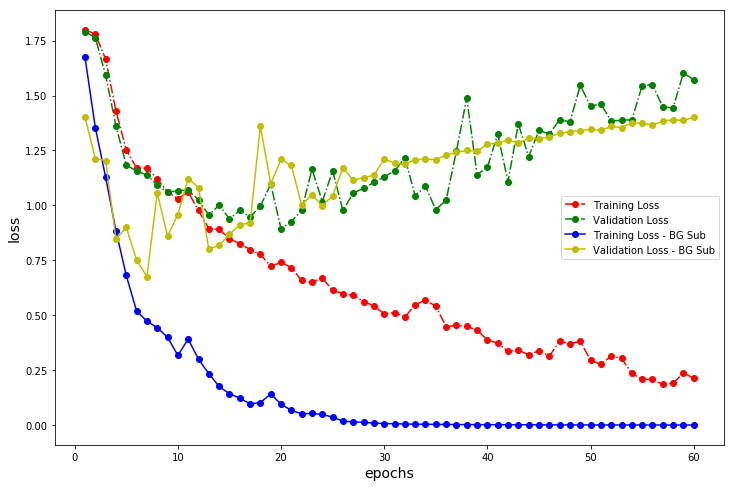}}
	\caption{Training and Validation loss for Model 2 using KTH Dataset}
	\label{KTH_M2}
	\vspace{-0.5em}
\end{figure}

\begin{figure}[htbp]
	\vspace{-1em}
	\centerline{\includegraphics[height=60mm]{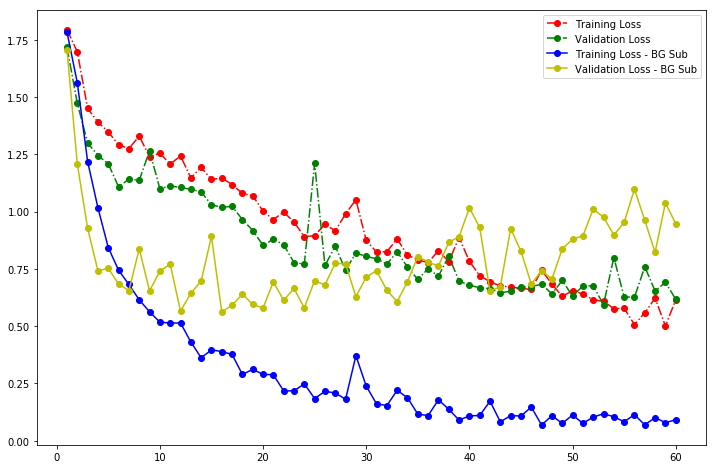}}
	\caption{Training and Validation loss for Model 3 using KTH Dataset}
	\label{KTH_M3}
	\vspace{-0.5em}
\end{figure}

\begin{figure}[htbp]
	\vspace{-1em}
	\centerline{\includegraphics[height=70mm]{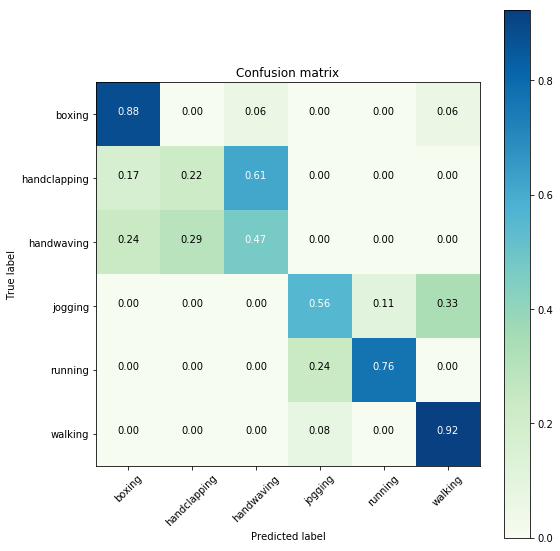}}
	\caption{Confusion matrix for Model 3 using KTH Dataset without background subtraction}
	\label{KTH_CM}
	\vspace{-0.5em}
\end{figure}

\begin{figure}[htbp]
	\vspace{-1em}
	\centerline{\includegraphics[height=70mm]{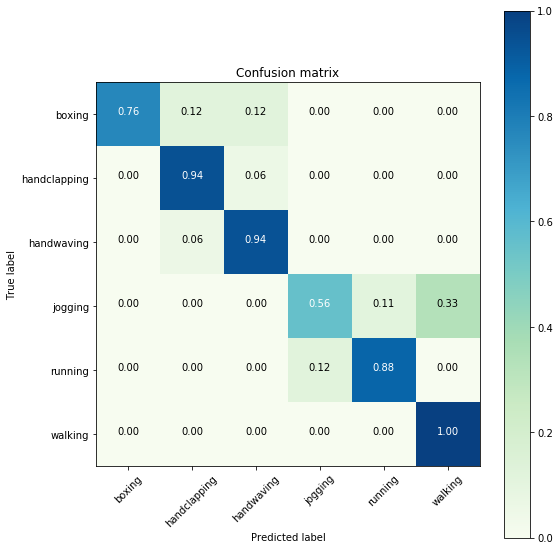}}
	\caption{Confusion matrix for Model 3 using KTH Dataset with background subtraction}
	\label{KTH_BG_CM}
	\vspace{-1em}
\end{figure}

\begin{figure}[htbp]
	\vspace{-0.5em}
	\centerline{\includegraphics[height=70mm]{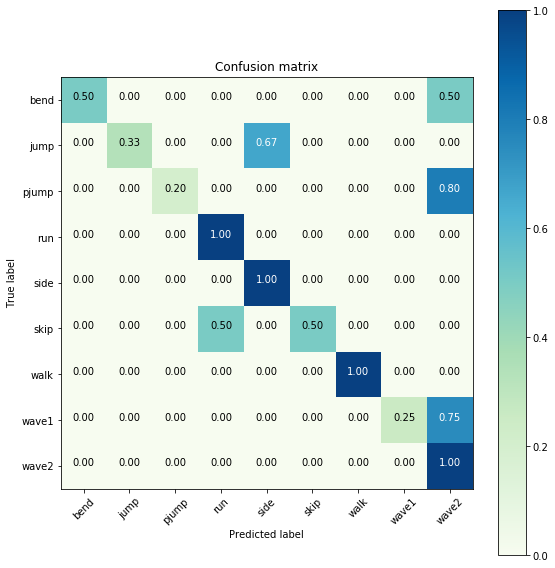}}
	\caption{Confusion matrix for Model 3 using Wiezmann Dataset without background subtraction}
	\label{W_CM}
	\vspace{-0.5em}
\end{figure}

\begin{figure}[htbp]
	\vspace{-0.5em}
	\centerline{\includegraphics[height=70mm]{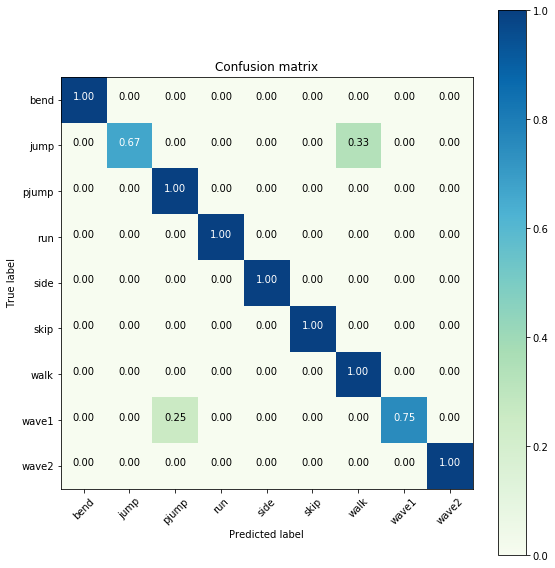}}
	\caption{Confusion matrix for Model 3 using Wiezmann Dataset with background subtraction}
	\label{W_BG_CM}
	\vspace{-0.5em}
\end{figure}

\begin{figure}[htbp]
	\vspace{-0.5em}
	\centerline{\includegraphics[height=70mm]{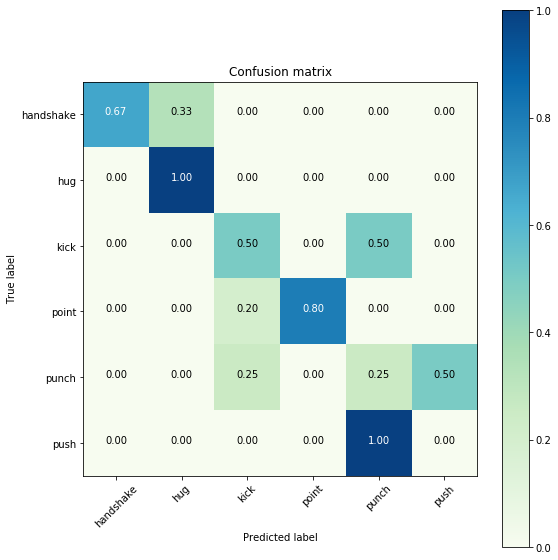}}
	\caption{Confusion matrix for Model 3 using UT-Interaction Dataset without background subtraction}
	\label{UT_CM}
	\vspace{-0.5em}
\end{figure}

\begin{figure}[htbp]
	\vspace{-0.5em}
	\centerline{\includegraphics[height=70mm]{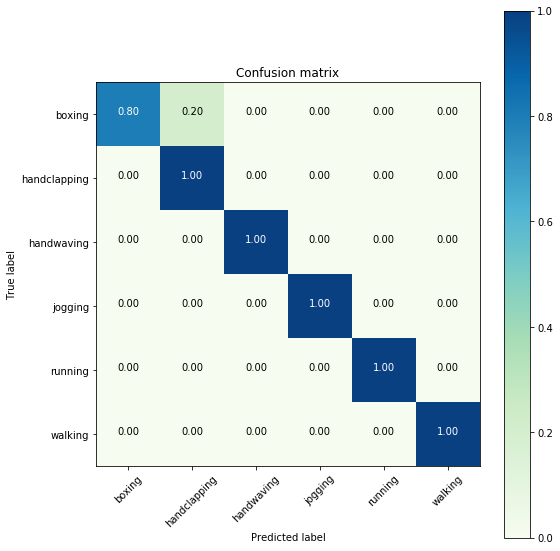}}
	\caption{Confusion matrix for Model 4 using UT-Interaction Dataset with background subtraction}
	\label{UT_BG_CM}
	\vspace{-0.5em}
\end{figure}

Following figures depicts graphical representation of results obtained in various scenarios.\\
Figures \ref{KTH_M1}-\ref{KTH_M3} are the results obtained on KTH Dataset:
The label \textbf{‘Training loss’} indicates the training loss for the videos without background subtraction.\\
The label \textbf{‘Validation loss’} indicates the validation loss for the videos without background subtraction.\\
The label \textbf{‘Training loss – BG Sub’} indicates the training loss for the videos with  background subtraction.\\
The label \textbf{‘Validation loss – BG Sub’} indicates the validation loss for the videos with background subtraction.\\
Figures \ref{KTH_CM}, \ref{KTH_BG_CM}, \ref{W_CM}, \ref{W_BG_CM}, \ref{UT_CM}, and \ref{UT_BG_CM} represents the confusion matrices obtained using model 3 on KTH, Weizmann and UT-Interaction data-sets respectively for with and without background subtraction.\\
\begin{figure}[htbp]
	\vspace{-1em}
	\centerline{\includegraphics[height=60mm]{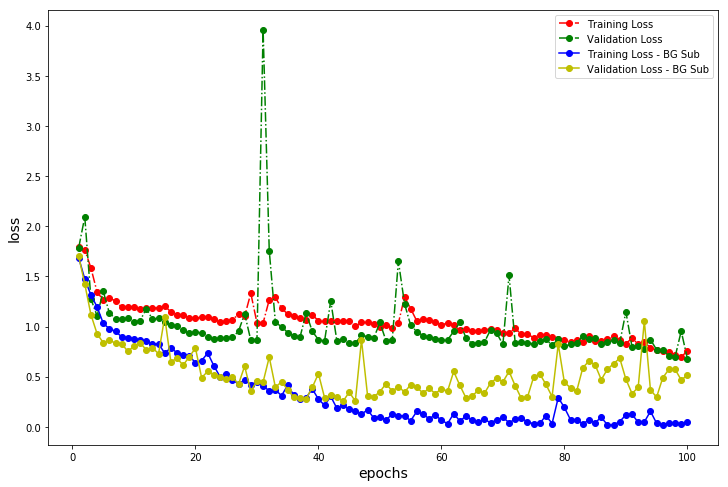}}
	\caption{Training and Validation loss for Model 4 using KTH Dataset}
	\label{KTH_M4}
	\vspace{-0.5em}
\end{figure}
\begin{figure}[htbp]
	\vspace{-1em}
	\centerline{\includegraphics[height=70mm]{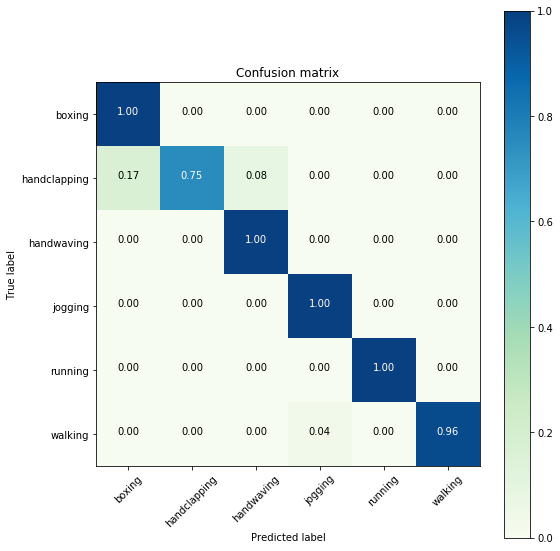}}
	\caption{Confusion matrix for Model 4 using KTH Dataset with background subtraction}
	\label{KTH4_BG_CM}
	\vspace{-0.5em}
\end{figure}
%%%%%%%%%%%%%% Weizmannn Dataset
\begin{figure}[htpb]
	\vspace{-1.5em}
	\centerline{\includegraphics[height=60mm]{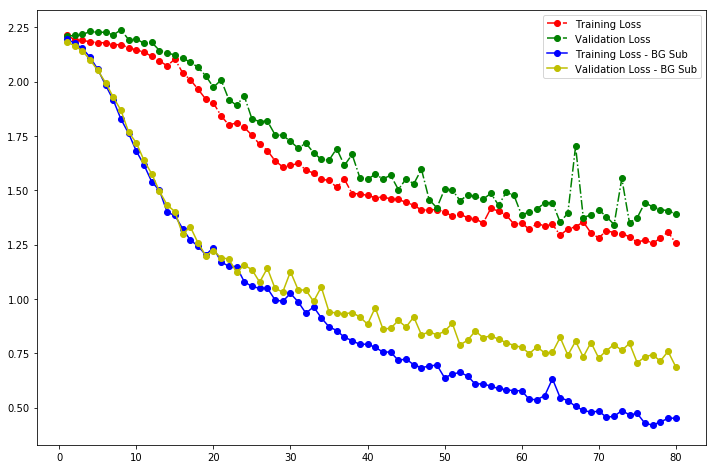}}
	\caption{Training and Validation loss for Model 1 using Wiezmann Dataset}
	\label{W_M1}
	\vspace{-0.5em}
\end{figure}
\begin{figure}[htbp]
	\vspace{-1em}
	\centerline{\includegraphics[height=60mm]{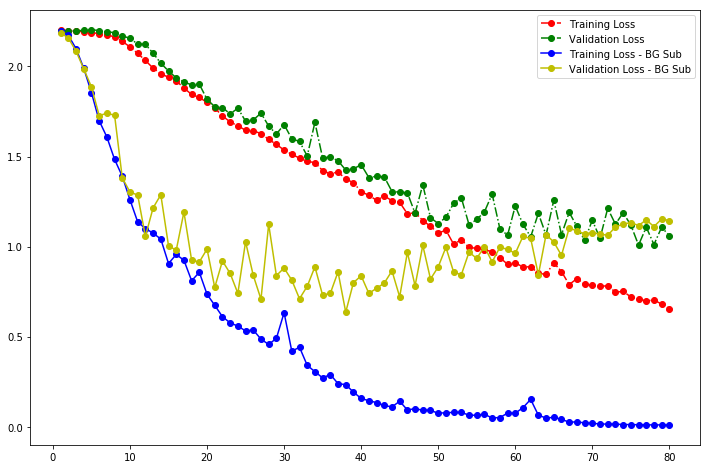}}
	\caption{Training and Validation loss for Model 2 using Wiezmann Dataset}
	\label{W_M2}
	\vspace{-0.5em}
\end{figure}
\begin{figure}[htbp]
	\vspace{-1em}
	\centerline{\includegraphics[height=60mm]{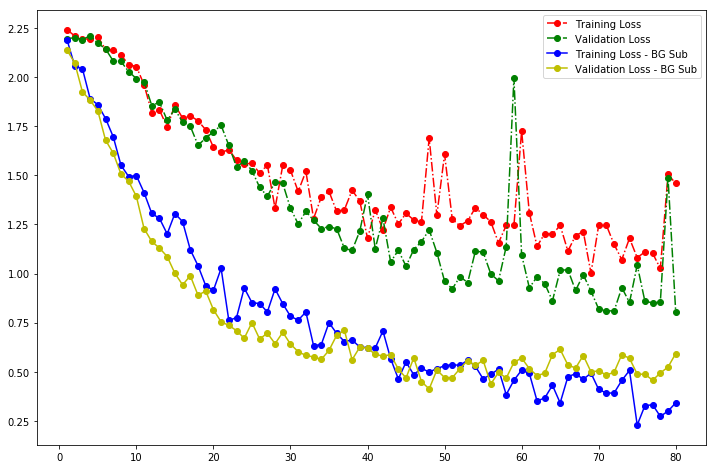}}
	\caption{Training and Validation loss for Model 3 using Wiezmann Dataset}
	\label{W_M3}
	\vspace{-0.5em}
\end{figure}
\begin{figure}[htbp]
	\vspace{-1em}
	\centerline{\includegraphics[height=60mm]{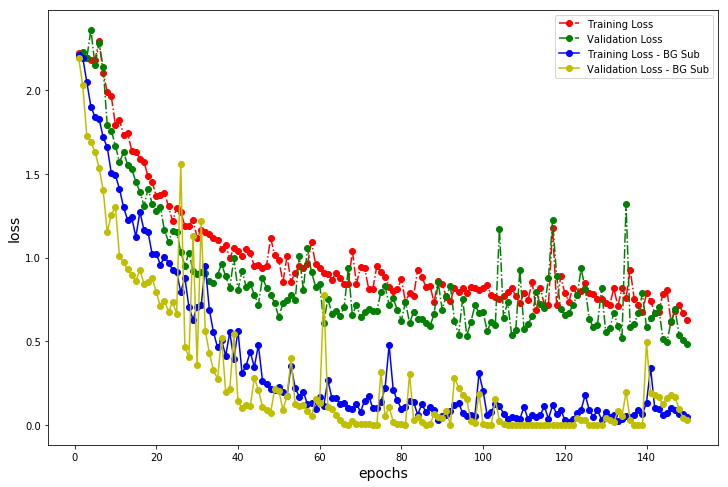}}
	\caption{Training and Validation loss for Model 4 using Wiezmann Dataset}
	\label{W_M4}
	\vspace{-0.5em}
\end{figure}
\begin{figure}[htbp]
	\vspace{-0.5em}
	\centerline{\includegraphics[height=70mm]{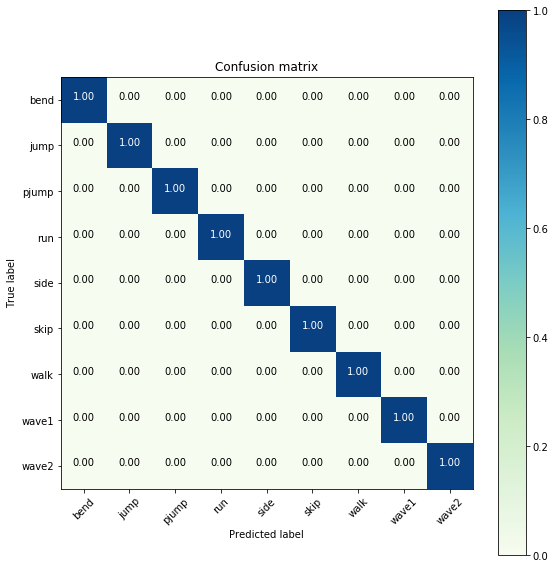}}
	\caption{Confusion matrix for Model 4 using Wiezmann Dataset with background subtraction}
	\label{W4_BG_CM}
	\vspace{-0.5em}
\end{figure}
%%%%%%%%%%%%%% UT-Interaction Dataset%%%%%%%%%%%%%%%%
\begin{figure}[htpb]
	\vspace{-1.5em}
	\centerline{\includegraphics[height=60mm]{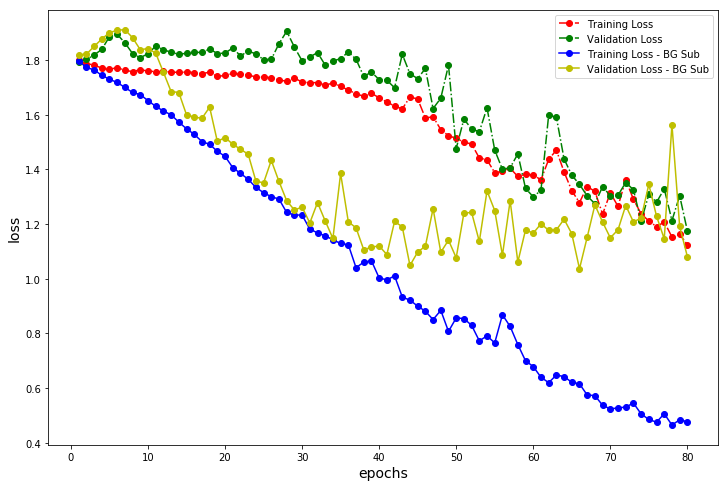}}
	\caption{Training and Validation loss for Model 1 using UT-Interaction Dataset}
	\label{UT_M1}
	\vspace{-0.5em}
\end{figure}
\begin{figure}[htbp]
	\vspace{-1em}
	\centerline{\includegraphics[height=60mm]{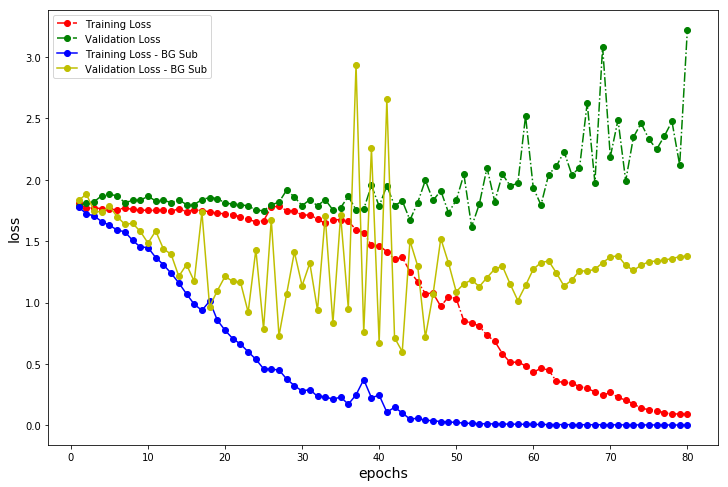}}
	\caption{Training and Validation loss for Model 2 using UT-Interaction Dataset}
	\label{UT_M2}
	\vspace{-0.5em}
\end{figure}
\begin{figure}[htbp]
	\vspace{-1em}
	\centerline{\includegraphics[height=60mm]{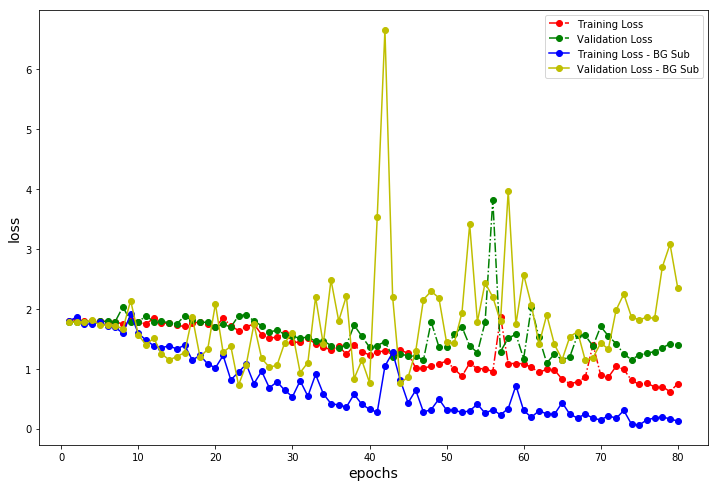}}
	\caption{Training and Validation loss for Model 3 using UT-Interaction Dataset}
	\label{UT_M3}
	\vspace{-0.5em}
\end{figure}
\begin{figure}[htbp]
	\vspace{-1em}
	\centerline{\includegraphics[height=60mm]{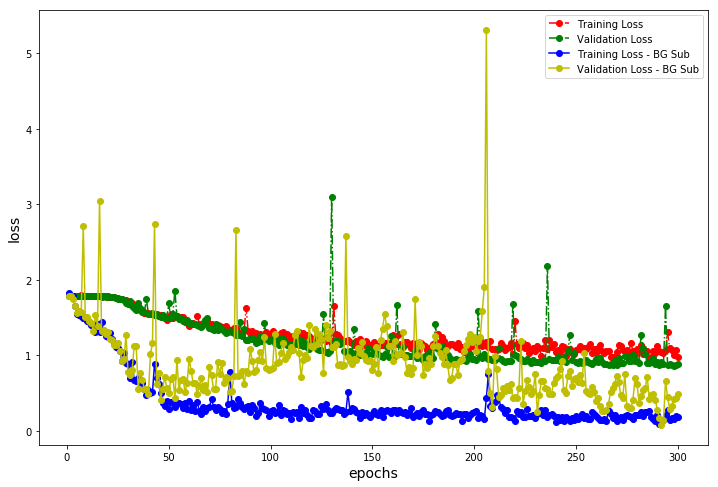}}
	\caption{Training and Validation loss for Model 4 using UT-Interaction Dataset}
	\label{UT_M4}
	\vspace{-0.5em}
\end{figure}
\begin{figure}[htbp]
	\vspace{-0.5em}
	\centerline{\includegraphics[height=70mm]{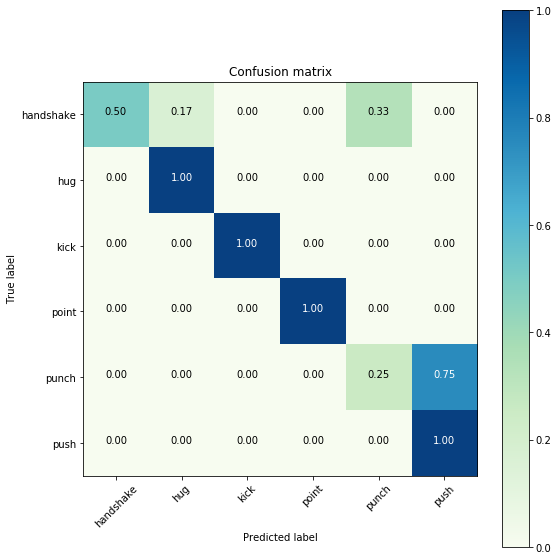}}
	\caption{Confusion matrix for Model 3 using UT-Interaction Dataset with background subtraction}
	\label{UT4_BG_CM}
	\vspace{-0.5em}
\end{figure}
\begin{figure*}[t]
	\vspace{-0.5em}
	\centerline{\includegraphics[height=120mm]{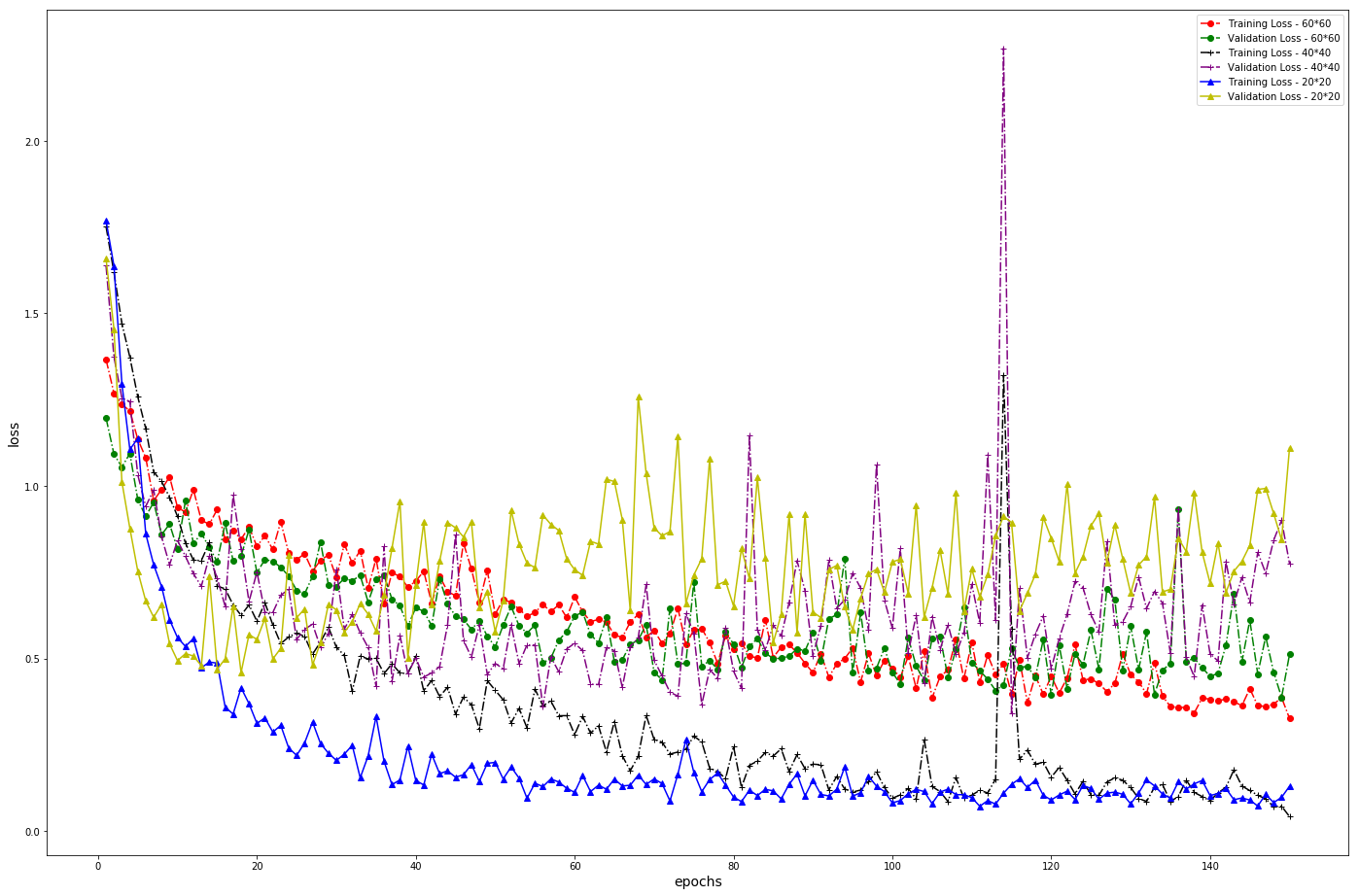}}
	\caption{Confusion matrix for Model 3 using UT-Interaction Data-set with background subtraction}
	\label{Size_vary}
	\vspace{-0.5em}
\end{figure*}
From the figures \ref{KTH_M1}-\ref{KTH_M4}, \ref{W_M1}-\ref{W_M4} and \ref{UT_M1}-\ref{UT_M4}, it can be observed that the model converges faster to minima with lesser number of epochs when the data used for training is the one with background subtraction. Also in many cases, the accuracy has improved when the background subtraction is performed. This shows that using the data that corresponds to the motion of the activity, improves the accuracy than using the data directly. However, it can also be observed that the model begins over-fitting in the earlier stages when the background subtraction is used.\\
Also, KTH and Weizmann data-sets corresponds to videos containing action, whereas UT-Interaction data-set contains the interaction between the 2 persons. From the above results, it can be observed that using background subtraction on interaction data-set performed better than not using the background subtraction as pre-processing for the videos. Also, the use of 3D Convolution Neural Network performed well on action as well as interaction recognition. Whereas, the state-of-art techniques used for implementing action recognition is different from that of techniques used for implementing interaction recognition.\\
From Model4 used on KTH, Weizmann and UT-Interaction data-sets respectively, it can be observed that data augmentation improves training of the model. Thus, improving the accuracy achieved.\\
\indent Tables \ref{KTH_Accuracy}, \ref{W_Accuracy} and \ref{US_Accuracy} depicts the accuracy obtained on KTH, Weizmann and UT-Interaction datasets respectively.
For the purpose of this implementation, size of the image frame of the video was varied from 20*20, 40*40 and 60*60 respectively on model 3 on KTH Data-set, with background subtraction performed on the videos. Same data in the training, testing and validation videos were considered. Figure \ref{Size_vary} depicts the graphical representation of training and validation loss, upon varying the frame resolution.\\
Table \ref{Size_Accuracy} shows results of accuracy obtained by varying size of the image frame.\\
\begin{table}[ht]
	\caption{Maximum accuracy Obtained on KTH Data-set}
	\begin{tabular}{|>{\centering\arraybackslash}p{1.5cm}|p{3cm}|p{3cm}|}
		\hline
		\textbf{Model}&\textbf{Accuracy (Without Background Subtraction)}&\textbf{Accuracy (With Background Subtraction)}\\ \hline
		Model 1 & 32.00\% & 67.00\% \\ \hline
		Model 2 & 57.00\% & 64.00\% \\ \hline
		Model 3 & 62.00\% & 84.00\% \\ \hline
		Model 4 & 73.00\% & 96.00\% \\ \hline
	\end{tabular}
	\label{KTH_Accuracy}
\end{table}

\begin{table}[ht]
	\caption{Maximum accuracy Obtained on Weizmann Data-set}
	\begin{tabular}{| >{\centering\arraybackslash}p{1.5cm} |p{3cm}|p{3cm}|}
		\hline
		\textbf{Model}&\textbf{Accuracy (Without Background Subtraction)}&\textbf{Accuracy (With Background Subtraction)}\\ \hline
		Model 1 & 26.00\% & 83.33\% \\ \hline
		Model 2 & 63.33\% & 86.67\% \\ \hline
		Model 3 & 63.33\% & 93.33\% \\ \hline
		Model 4 & 80.00\% & 100.00\% \\ \hline
	\end{tabular}
	\label{W_Accuracy}
\end{table}
\begin{table}[ht]
	\caption{Maximum accuracy Obtained on UT-Interaction Data-set}
	\begin{tabular}{|>{\centering\arraybackslash}p{1.5cm}|p{3cm}|p{3cm}|}
		\hline
		\textbf{Model}&\textbf{Accuracy (Without Background Subtraction)}&\textbf{Accuracy (With Background Subtraction)}\\ \hline
		Model 1 & 45.00\% & 70.00\% \\ \hline
		Model 2 & 50.00\% & 75.00\% \\ \hline
		Model 3 & 60.00\% & 70.00\% \\ \hline
		Model 4 & 60.00\% & 80.00\% \\ \hline
	\end{tabular}
	\label{US_Accuracy}
\end{table}
\begin{table}[htpb]
	\caption{Accuracy obtained on varying frame size of videos using Model 3}
	\begin{tabular}{|>{\centering\arraybackslash}p{4cm}|>{\centering\arraybackslash}p{4cm}|}
		\hline
		\textbf{Frame size}&\textbf{Accuracy}\\ \hline
		20*20 & 84.00\% \\ \hline
		40*40 & 82.00\% \\ \hline
		60*60 & 84.00\% \\ \hline
	\end{tabular}
	\label{Size_Accuracy}
\end{table}
\section{IoT Framework for Activity Recognition}
\begin{figure*}[htbp]
	\centerline{\includegraphics[height=90mm]{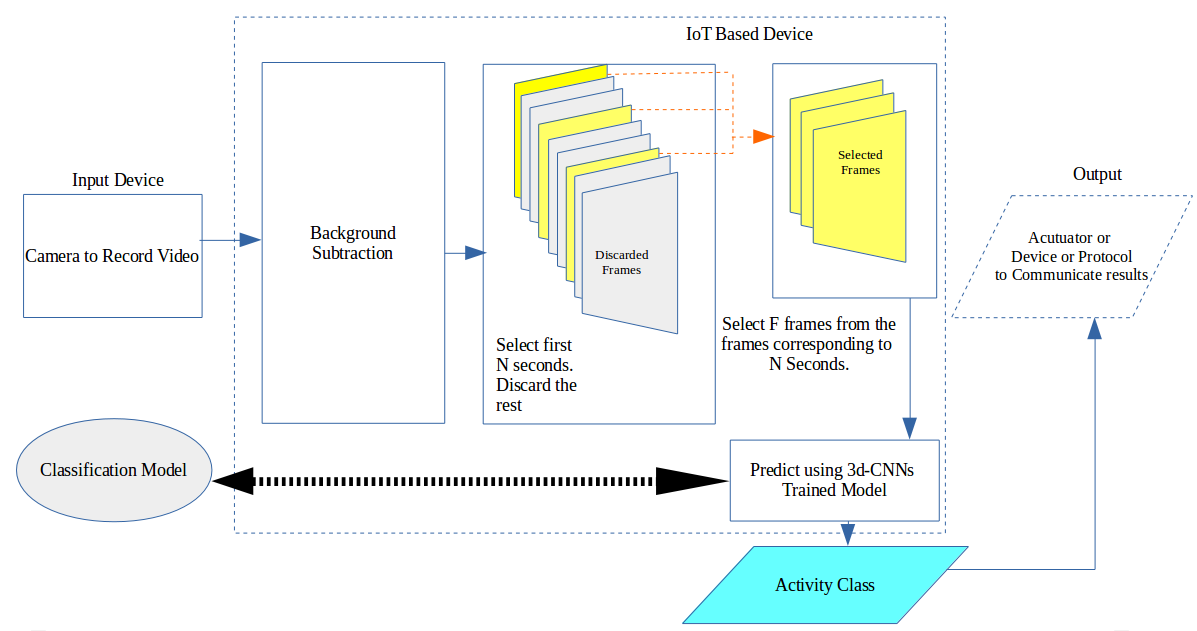}}
	\caption{ IoT Based Framework for Activity Recognition}
	\label{IOT}
\end{figure*}

\indent \indent The above system has been extended by implementing on Raspberry Pi, which acts as a mobile device and improves the portability of the device. Figure \ref{IOT} shows the architecture for implementing activity recognition in Raspberry Pi along with framework for IoT applications.
\\
\\
Following is the brief description of the IoT based framework for activity recognition:
\begin{itemize}
	\item The videos of the actions are recorded using input device, a camera module. A specific length of video of ‘S’ seconds is recorded using this device.
	\item The input is then processed, where the background subtraction is applied over the input video. Further, ‘N’ frames from the first ‘S’ seconds are considered by equal interleaving between the frames. This module for pre-processing is same as the regular implementation of pre-processing discussed in the previous sections. In our case, the pre-processing is done on raspberry-pi device itself.
	\item The model is trained on the actual test-bench and not on raspberry-Pi due to its limited computational capabilities. The trained classification model is loaded into raspberry-pi using model weights trained on our test bench.
	\item For the purpose of evaluation of the trained model, prediction of the activity class is made from the video recorded from raspberry-pi.
	\item The results of the classification can be further used to actuate an actuator or the results can be communicated to the concerned authority using communication devices or protocols. 
\end{itemize}
Below is the brief discussion on implementation of activity recognition on IoT Device:\\
The training was done using KTH data-set, using the system with given specification. The model weights after training was used in IoT based device, i.e, Raspberry Pi. This was done because low computational capabilities of Raspberry Pi, which would take exorbitant duration to train the model.\\
The video using the camera module of Raspberry Pi has been recorded at resolution of 160x120, which is similar to the size of video in the data-set. The video is for first 8 seconds, from which 35 frames from first 7 seconds is considered. This ensures that the size of input data is similar to size of the data used for training the system. The video was captured in an indoor environment with static camera as well as static background. The activities performed were same as the ones that are available in KTH dataset. These activities resulted in correct prediction, thus, leveraging the application of activity recognition system to portable environment. The results obtained here can be communicated using various methods, such as use of GSM module, to communicate results via SMS, or use of SMTP protocol to send an email alert, etc. The results obtained by recognizing the activities through such devices can be used for various purpose, such as alerting the concerned authority in surveillance environment, actuating the device in an IoT application, etc.
\section{Conclusion}
Video based human activity recognition system is used in many modern applications, such as monitoring system and those systems that need to respond to the activities of the person, such as Human Computer Interface systems. These systems are only capable of recognizing either the actions or the interactions. A novel approach for activity recognition based on background subtraction for the videos prior to the use of 3D-CNNs, which is suitable for both action and interaction data-set has been implemented. It is observed that use of background subtraction as the pre-processing technique yielded better results than using the video frames directly for the 3D-CNNs. The study of varying frame resolution of the video resulted in the findings that use of lower resolution of videos can train the system faster and are computationally cheaper when compared with the videos with higher resolution. Thus, it can be concluded that, though Convolutional Neural Networks do not require any pre-processing, value addition to the system in the case of human activity recognition system with static background environment is evidenced due to this pre-processing on the input data. The accuracy of the model can further be enhanced by reducing the problem of over-fitting and using better regularization. Other Deep learning approaches such as RNN and LSTMs can be used instead of CNNs. The given implementation works when there is no motion in the background or when there is no motion in the camera. Such scenarios can be considered in the future.\\
In the second stage of implementation, an IoT framework using Raspberry Pi was implemented to leverage the activity recognition system to portable devices, which was tested by recording own set of test data. The outcomes of recording the video using Raspberry Pi camera module and performing the classification of activities led to correct prediction of activity class. This leverages portability and networking capabilities of system, since wide range of interfacing options are available for Raspberry Pi as an IoT enabling device. However, it is observed that processing of video using Raspberry Pi is slower than the system used in the first stage, which is due to the low computational capabilities, that can overcome in future with advancement in hardware technologies.\\

\section*{Acknowledgment}
The work reported in this paper is supported by the college [BMSCE, Bengaluru] through the TECHNICAL EDUCATION QUALITY IMPROVEMENT PROGRAMME [TEQIP-III] of the MHRD, Government of India.

\bibliographystyle{IEEEtran}
\bibliography{IEEEabrv,AR_project_combined_springer}
\end{document}